\documentclass[letterpaper, 10 pt, conference]{ieeeconf}  

\usepackage[ansinew]{inputenc}
\usepackage{float}
\usepackage{graphicx}
\usepackage{fancyhdr}
\usepackage{amsmath,amssymb}
\usepackage{color}
\usepackage{verbatim}
\usepackage{scrtime}
\usepackage{float}
\usepackage[sort]{cite}
\usepackage{subfigure}
\usepackage{url}
\usepackage{multirow}
\usepackage[usenames,dvipsnames]{xcolor}
\usepackage{hyperref}
\hypersetup{
     colorlinks   = true,
     citecolor    = blue
}
\hypersetup{linkcolor=blue}
\usepackage{capt-of}

\newcommand{\changeBM}[1]{#1}

\newcommand{\rank}{\mathrm{rank}}
\newcommand{\trace}{{\rm Trace}}

\newcommand{\GL}[1]{\mathrm{GL}({#1})}

\newcommand{\OS}{\mathrm{OS}}

\newcommand{\mat}[1]{{\bf #1}}

\IEEEoverridecommandlockouts                              
\usepackage{arydshln}
\makeatletter
\def\hlinewd#1{%
  \noalign{\ifnum0=`}\fi\hrule \@height #1 \futurelet
   \reserved@a\@xhline}
\makeatother

\title{\LARGE \bf
Scaled stochastic gradient descent for \\low-rank matrix completion}

\author{Bamdev Mishra$^\dagger$\thanks{$^\dagger$Amazon Development Centre India, Bangalore 560055, India (Bamdevm@amazon.com). Initial work was done while this author was with the Department of Electrical Engineering and Computer Science, University of Li\`ege, 4000 Li\`ege, Belgium and was visiting the Department of Engineering (Control Group), University of Cambridge, Cambridge CB2 1PZ, UK.}
 and Rodolphe Sepulchre$^{\S}$\thanks{$^\S$University of Cambridge, Department of Engineering, Trumpington Street, Cambridge CB2 1PZ, UK (R.Sepulchre@eng.cam.ac.uk).}
}

\begin{document}

\maketitle
\thispagestyle{empty}
\pagestyle{empty}

\begin{abstract}
The paper looks at a scaled variant of the stochastic gradient descent algorithm for the matrix completion problem. Specifically, we propose a novel matrix-scaling of the partial derivatives that acts as an efficient preconditioning for the standard stochastic gradient descent algorithm. This proposed matrix-scaling provides a trade-off between local and global second order information. It also resolves the issue of scale invariance that exists in matrix factorization models. The overall computational complexity is linear with the number of known entries, thereby extending to a large-scale setup. Numerical comparisons show that the proposed algorithm competes favorably with state-of-the-art algorithms on various different benchmarks.
\end{abstract}

\section{Introduction}
The problem of low-rank matrix completion amounts to completing a matrix from a small number of entries by assuming a low-rank model for the matrix. This problem has been addressed both from theoretical \cite{candes09b, keshavan10a, wei15a} as well as from algorithmic viewpoints \cite{cai10a, keshavan10a, keshavan09a, boumal15a, recht10a, yu14a, tanner15a, vandereycken13a, ngo12a, recht13a, teflioudi12a, shalit10a, balzano10a,  zhou08a, pilaszy10a, wen12a}. A standard way of approaching the problem is by casting it as a \emph{fixed-rank optimization problem} with the assumption that the optimal rank $r$ is known a priori, i.e.,
\begin{equation}\label{eq:matrix-completion-formulation}
\begin{array}{llll}
	\min\limits_{\mat{X}\in\mathbb{R}^{n \times m}}
		&	\frac{ 1}{2}\|\mathcal{P}_{\Omega}(\mat{X}) - \mathcal{P}_{\Omega}(\mat{X^\star})\|_F^2 \\
	 \text{subject to} & \rank(\mat{X})=r,
\end{array}
\end{equation}
where ${\mat{X ^ \star}}\in\mathbb{R}^{n\times m}$ is a matrix whose entries are known for indices if they belong to the subset $(i,j)\in\Omega$ and $\Omega$ is a subset of the complete set of indices $\{(i,j):i\in\{1,...,n\}\text{ and }j\in\{1,...,m\}\}$. The operator $\mathcal{P}_{\Omega}(\mat{X}_{ij})=\mat{X}_{ij}$ if $(i,j) \in \Omega$ and $\mathcal{P}_{\Omega}(\mat{X}_{ij})=0$ otherwise is called the orthogonal sampling operator and is a mathematically convenient way to represent the subset of entries. $\|\cdot \|_F$ is the \emph{Frobenius} norm and $|\Omega|$ is the number of known entries. The low-rank assumption on (\ref{eq:matrix-completion-formulation}) implies that $r \ll \min(n, m )$. The rank constraint correlates the known entries with the unknown ones. Recent contributions provide bounds on $|\Omega|$ (linear in $n$ and $m$) for which exact reconstruction is possible in certain conditions from entries sampled randomly \cite{candes09b,keshavan10a}. 

Problem (\ref{eq:matrix-completion-formulation}) and its (many) variants find applications in control systems and system identification \cite{markovsky13a}, machine learning \cite{recht13a}, and information theory \cite{shi16a}, to name a just few. A popular way to tackle the rank constraint in (\ref{eq:matrix-completion-formulation}) is by using a factorization model. The earlier works \cite{meyer11a, mishra12a} discuss factorization models and show how to perform first and second order optimization with them in presence of \emph{scale invariance}, which arises due to non-uniqueness of factorization models. \cite{vandereycken13a, shalit10a} exploit the \emph{Riemannian} structure of rank constraint and provide a spectrum of algorithms. \emph{Preconditioning} with rank constraint in the context of matrix completion is recently explored in \cite{boumal15a, ngo12a, mishra12a, mishra16a}. Alternating minimization algorithms that exploit the \emph{least-squares structure} of the cost function of (\ref{eq:matrix-completion-formulation}) are proposed in \cite{wen12a,tanner15a}. The least-squares structure is also exploited in \cite{keshavan10a, boumal15a} to develop algorithms on the \emph{Grassmann} manifold. The Matlab toolbox Manopt contains various other implementations \cite{boumal14a}.

While all the earlier mentioned algorithms are sequential algorithms, the works \cite{yu14a, recht13a, teflioudi12a, zhou08a, pilaszy10a} focus on parallel and stochastic versions.  An alternating least-squares approach is proposed in \cite{zhou08a} to learn the \emph{rows} of a factorization model, where each subproblem has a closed-form solution. The paper \cite{pilaszy10a} also exploits the least-squares structure, but at the level of the entries of the rows of factorization models. \cite{yu14a} focuses on learning $r$ rank-$1$ factorizations cyclically, where each subproblem is solved using the algorithm of \cite{pilaszy10a}. The stochastic gradient descent algorithm (SGD) proposed in \cite{recht13a} updates the factorization model as and when the known entries are observed. The specific focus there is on parallelization of the SGD algorithm. A distributed version of SGD is proposed in \cite{teflioudi12a}. Another approach that is suitable in an online setup is proposed  in \cite{balzano10a}, where the data is assumed to be streaming from low-dimensional subspaces. The works \cite{shalit10a, meyer11a} exploit the Riemannian structure of rank constraint to propose online algorithms for low-rank matrix completion. 

Our focus in this paper is on a \emph{scaled} variant of SGD that accelerates the standard SGD algorithm and respects the scale invariance property of the factorization model. To achieve this, we propose a novel matrix-scaling of the partial derivatives in Section \ref{sec:proposed_sgd} that combines \emph{global and local} second order information. The computational cost of the algorithm per pass through $|\Omega|$ known entries is $O(|\Omega| (r^3/ b + b_{\mat L}r^2/b +  b_{\mat R}r^2/b  + r + {\rm log}b) )$, where $b$ is the batch size of the entries that we pick and $b_{\mat L}$ and $b_{\mat R}$ are the rows of $\mat L$ and $\mat R$ that are updated. The computational cost is comparable to those of \cite{zhou08a, recht13a, yu14a} for $r \ll {\rm min}(n, m)$. Our numerical comparisons in Section \ref{sec:numerical_comparisons} suggest that the proposed scaled SGD algorithm competes favorably with state-of-the-art on a number of different benchmarks, especially outperforming others on ill-conditioned and scarcely sampled data.

\section{Scaled stochastic gradient descent}\label{sec:proposed_sgd}

\begin{table}[t]
\caption{{Proposed stochastic gradient descent algorithm for (\ref{eq:matrix-completion-formulation}).}}
\label{tab:Riemannian_stochastic_gradient_descent} 
\begin{center} 
\begin{tabular}{ |p{8.3cm}| }
\hline
\begin{enumerate}
\item Pick $b$ known entries with their indices.
\item Set up the completion subproblem by finding the indices corresponding to the submatrices $\mat{L}_b$ and $\mat{R}_b$, which need to be modified. Consequently, find the subset $\Omega_b$ of indices out of the total $b_{\mat L} b_{\mat R}$ indices.

\item Compute the residual $\mat{S}_b = \mathcal{P}_{{\Omega}_b}( \mat{L}_b \mat{R}_b^T -  \mat{X}_b ^\star)$.

\item Given a stepsize $t$, update $\mat{L}_b$ and $\mat{R}_b$ as 
\[
\begin{array}{lllll}
{\mat{L}_b}_+  = {\mat{L}_b} \\
\qquad \qquad -  t \mat{S}_b \mat{R}_b (\frac{b\mu}{\max(m, n)}(\mat{R}^T\mat{R}) + (1 -\mu) (\mat{R}_b^T\mat{R}_b) ) ^{-1}\\
{\mat{R}_b}_+  =  {\mat{R}_b} \\
\qquad \qquad  -  t \mat{S}_b^T \mat{L}_b  (\frac{b\mu}{\max(m, n)}(\mat{L}^T\mat{L}) + (1-\mu) (\mat{L}_b^T\mat{L}_b) ) ^{-1}.\\
\end{array}
\]
\item Update $\mat{L}^T\mat{L}$ and $\mat{R}^T\mat{R}$.
\item Repeat.
\end{enumerate}
  \\
 \hline
\end{tabular}
\end{center} 
\end{table}

Given a matrix $\mat{X}$ of size ${n \times m}$ and rank $r$, it admits the factorization
\begin{equation}\label{eq:factorization}
\mat{X} = \mat{L} \mat{R}^T,
\end{equation}
where $\mat{L} \in \mathbb{R}_* ^{n \times r}$ and $\mat{R} \in \mathbb{R}^{m \times r}_*$, \changeBM{where $ \mathbb{R}_* ^{n \times r}$ is the set of $n\times r$} full column-rank matrices \cite{piziak99a}. Consequently, the problem (\ref{eq:matrix-completion-formulation}) boils down to 
\begin{equation}\label{eq:matrix-completion-formulation_factored}
\begin{array}{llll}
	\min\limits_{\mat{L}\in\mathbb{R}_*^{n \times r}, \mat{R}\in\mathbb{R}_*^{m \times r}}
		&	\frac{ 1}{2}\|\mathcal{P}_{\Omega}(\mat{LR}^T) - \mathcal{P}_{\Omega}(\mat{X^\star})\|_F^2.
\end{array}
\end{equation}

Consider a stochastic gradient setup for solving (\ref{eq:matrix-completion-formulation_factored}), where we pick $b$ known entries at a time and then take a gradient descent step that updates the matrices $\mat{L}$ and $\mat{R}$. Due to the cost function structure, we end up updating only a \emph{maximum of} $b$ rows of $\mat{L}$ and $\mat{R}$ at a time. Let $b_{\mat L}$ rows of $\mat{L}$ and $b_{\mat R}$ rows of $\mat R$ be updated when $b$ known entries are picked, where $b_{\mat L} \leq b$ and $b_{\mat R} \leq b$. Let $\mat{L}_b$ be the corresponding submatrix of $\mat{L}$ with the $b_{\mat L}$ rows, i.e., its size is $b_{\mat L} \times r$. Similarly, let $\mat{R}_b$ be the submatrix of $\mat{R}$ with the $b_\mat{R}$ rows and size $b_{\mat R} \times r$. An interpretation is that, each time we pick $b$ known entries, we have a \changeBM{subproblem of completing} a matrix $\mat{X}_b^\star$ of size $b_{\mat L} \times b_\mat{R}$ with $b$ known entries at indices $\Omega_b$, which needs to be approximated by $\mat{L}_b \mat{R}_b ^T$. If $\mat{S}_b$ is the residual matrix of this subproblem, then the partial derivatives at $(\mat{L}_b, \mat{R}_b)$ are $(\mat{S}_b \mat{R}_b,  \mat{S}_b^T \mat{L}_b)$, where $\mat{S}_b = \mathcal{P}_{{\Omega}_b}( \mat{L}_b \mat{R}_b^T -  \mat{X}_b ^\star)$ is of size $b_{\mat L} \times b_\mat{R}$.

The proposed stochastic gradient descent updates are
\begin{equation}\label{eq:modified_scaled_stochastic}
\begin{array}{lllll}
{\mat{L}_b}_+  =  {\mat{L}_b}  -  t \mat{S}_b \mat{R}_b (\frac{b\mu}{\max(m, n)}(\mat{R}^T\mat{R}) + (1 -\mu) (\mat{R}_b^T\mat{R}_b) ) ^{-1}\\
{\mat{R}_b}_+  =  {\mat{R}_b}  -  t \mat{S}_b^T \mat{L}_b  (\frac{b\mu}{\max(m, n)}(\mat{L}^T\mat{L}) + (1-\mu) (\mat{L}_b^T\mat{L}_b) ) ^{-1},\\
\end{array}
\end{equation} 
where $t$ is the step size, $b/\max(m, n)$ is a normalization constant, and $\mu$ is a nonnegative scalar in $[0, 1]$ that weighs $\mat{L}^T\mat{L}$ and $\mat{L}_b^T\mat{L}_b$ differently. \changeBM{The term $b/\max(m, n)$ ensures that the Frobenius norm of ${b}(\mat{L}^T\mat{L})/{\max(m, n)}$  and $ (\mat{L}_b^T\mat{L}_b)  $ are of the same order.} Similarly, the terms $\mat{R}^T\mat{R}$ and $\mat{R}_b^T\mat{R}_b$. \changeBM{It should be stated that $\mat{L}^T\mat{L}$ and $\mat{R}^T\mat{R}$ can be either be computed after every update or after a certain number of updates (e.g., one pass through data).} The stepsize $t$ can be modified, e.g., using the bold driver protocol \cite{battiti89a, teflioudi12a} or the exponential decay protocol \cite[Section~4.1]{recht13a}. The choice of $\mu$ depends on the problem.

The proposed algorithm is shown in Table \ref{tab:Riemannian_stochastic_gradient_descent}.

It should be noted that the terms $\mat{L}^T\mat{L}$ and $\mat{R}^T\mat{R}$ capture a part of second order information of $\|\mat{LR}^T - \mat{X^\star}\|_F^2/2$, which is related to the cost function of (\ref{eq:matrix-completion-formulation_factored}) and gives a simpler way to understand the behavior of the cost function. In fact, $\|\mat{LR}^T - \mat{X^\star}\|_F^2/2$ is obtained by assuming that $\Omega$ is the full set of indices in (\ref{eq:matrix-completion-formulation_factored}). This relies on \emph{strict convexity} of $\|\mat{LR}^T - \mat{X^\star}\|_F^2/2$ with \changeBM{respect to} the factors $\mat L$ and $\mat R$ individually. \changeBM{The block diagonal approximation of the Hessian of $\|\mat{LR}^T - \mat{X^\star}\|_F^2/2$ with respect to $(\mat{L}, \mat{R})$ is $((\mat{R}^T \mat{R}) \otimes \mat{I}_{n}, (\mat{L}^T \mat{L}) \otimes \mat{I}_{m})$, where $\mat{I}_n$ is the $n\times n$ identity matrix and $\otimes$ is the Kronecker product of matrices.} For example, this Hessian approximation is used in the works of \cite{boumal15a, ngo12a, mishra12a}, where the authors accelerate the convergence of algorithms (e.g., steepest descent, conjugate gradients, and trust-regions) by scaling the partial derivatives with respect to $\mat L$ and $\mat R$ with $(\mat{R}^T\mat{R})^{-1}$ and $(\mat{L}^T\mat{L})^{-1}$, respectively. 

In our case, an additional motivation for a trade-off with $\mu$ in (\ref{eq:modified_scaled_stochastic}), i.e., between the terms $\mat{L}^T\mat{L}$ and $\mat{R}^T\mat{R}$ on one hand and $\mat{L}_b^T\mat{L}_b$  and  $\mat{R}_b^T\mat{R}_b$ on the other, comes from the following intuition. $\mat{L}^T\mat{L}$ and $\mat{R}^T\mat{R}$ can be interpreted as capturing the global second order information as they contain knowledge of all the rows of $\mat{L}$ and $\mat{R}$ \cite{ngo12a, mishra12a, boumal15a}. On the other hand in a stochastic setup we modify only $\mat{L}_b$ rows of $\mat{L}$ and $\mat{R}_b$ rows of $\mat{R}$. This leads to the argument that $\mat{L}_b^T\mat{L}_b$ in the term $\mat{L}^T\mat{L}$ should be given a higher weight. Similarly, the part $\mat{R}_b^T\mat{R}_b$ in $\mat{R}^T\mat{R}$ is given a \emph{differentiated} weight. Overall the terms $(\frac{b\mu}{\max(m, n)}(\mat{R}^T\mat{R}) + (1 -\mu) (\mat{R}_b^T\mat{R}_b) ) ^{-1}$ and $(\frac{b\mu}{\max(m, n)}(\mat{L}^T\mat{L}) + (1-\mu) (\mat{L}_b^T\mat{L}_b) ) ^{-1}$ which are multiplied to the partial derivatives $\mat{S}_b \mat{R}_b$ and $  \mat{S}_b^T \mat{L}_b$, respectively, act as an \emph{efficient preconditioner} for the standard stochastic gradient descent updates proposed in \cite{recht13a, teflioudi12a}. 

\changeBM{Conceptually, our approach can also be connected to the recent work \cite{johnson13a}, which also combines local and global information, but in the context of \emph{first order} information, i.e., gradient information. The resulting stochastic variance reduction algorithms have shown superior performance.}

\textbf{A. Computation cost.}\label{sec:computational_cost}
In Table \ref{tab:Riemannian_stochastic_gradient_descent}, Step $2$ costs $O(b {\rm log}b)$ to identify the rows of $\mat{L}$ and $\mat{R}$ that need to be modified and costs $O(b)$ to find the subset $\Omega _b$. It involves \emph{sorting} \changeBM{(and hence, the cost is $O(b {\rm log}b)$)} the row indices corresponding to the $b$ known entries to find the unique rows of $\mat{L}$ and $\mat{R}$ that are required to be updated. Computation of the residual $\mat{S}_b$ costs $O(b)$ in Step $3$. The updates in Step $4$ costs $O(r^3 + br)$. Step $5$ costs $O(b_{\mat L} r^2 + b_{\mat{R}} r^2)$. Consequently, each time we pick any $b$ entries, our proposed gradient descent step costs $O(b_{\mat L} r^2 + b_{\mat{R}} r^2 + br+ b + b{\rm log} b + r^3)$, where $b_{\mat L} \leq b$ and $b_{\mat R} \leq b$. Equivalently, our algorithm costs $O(|\Omega| (r^3/ b + b_{\mat L}r^2/b +  b_{\mat R}r^2/b  + r + {\rm log}b) )$ after we have seen $| \Omega|$ entries.

Depending on $b$, the computational cost varies from $O(|\Omega| r^3)$ to $O(|\Omega| r^2)$. For $b = 1$, the inverse computation of $(\frac{b\mu}{\max(m, n)}(\mat{R}^T\mat{R}) + (1 -\mu) (\mat{R}_b^T\mat{R}_b) )$ costs $O(r^2)$ as it requires only a rank-$1$ modification per update. Consequently, our algorithm costs $O(|\Omega| r^2)$ for $b = 1$.  For $b \in (1, r )$, the computation cost is upper bounded by $O(|\Omega| r^3)$ for a straightforward implementation of matrix inversion in (\ref{eq:modified_scaled_stochastic}). For \changeBM{$b \geq r$}, the computational cost is $O(|\Omega| r^2)$. In particular if $b = |\Omega|$, then $b_{\mat L} = n$, $b_{\mat R} = m$, and the computational cost is $O(|\Omega |r + nr^2 + mr^2)$, which is same as the computational cost (per iteration) of most algorithms in the batch setup, e.g., the ones proposed in \cite{mishra12a, boumal15a, vandereycken13a,tanner15a,ngo12a,keshavan10a,wen12a}.

\textbf{B. Scale invariance.}
The proposed updates (\ref{eq:modified_scaled_stochastic}) also resolve the issue of scale invariance arising from non-uniqueness of matrix factorization (\ref{eq:factorization}) as $\mat{X}$ remains unchanged under the action \cite{piziak99a, meyer11a}
\begin{equation}\label{eq:symmetry}
(\mat{L},\mat{R})\mapsto (\mat{L}\mat{M}^{-1},\mat{R}\mat{M}^{T}),
\end{equation}
for all non-singular matrices $\mat{M} \in \GL{r}$, where $ \GL{r}$ is the set of all non-singular matrices of size $r \times r$. Equivalently, $\mat{X} = \mat{L} \mat{R}^T = \mat{L} \mat{M}^{-1}  (\mat{R}   \mat{M}^T)^T$. The issue of scale invariance refers to the behavior of algorithms which behave \emph{equivalently} when initialized, say, either with $(\mat{L}_0,\mat{R}_0)$ or with $(\mat{L}_0\mat{M}^{-1},\mat{R}_0\mat{M}^{T})$ for all non-singular matrices $\mat{M} \in \GL{r}$. For example, if $ ({\mat{L}}_+,{\mat{R}}_+)$ is an update obtained from $(\mat{L}_0,\mat{R}_0)$, then a \emph{scale-invariant algorithm} produces the update $({\mat{L}}_+\mat{M}^{-1},{\mat{R}}_+\mat{M}^{T})$ starting from $(\mat{L}_0\mat{M}^{-1},\mat{R}_0\mat{M}^{T})$. It is the case for the algorithm proposed in Table \ref{tab:Riemannian_stochastic_gradient_descent}. It is straightforward to show that under the mapping (\ref{eq:symmetry}), the proposed updates (\ref{eq:modified_scaled_stochastic}) lead to the transformation $ ({\mat{L}_b}_+,{\mat{R}_b}_+)\mapsto ({\mat{L}_b}_+\mat{M}^{-1},{\mat{R}_b}_+\mat{M}^{T})$. On the other hand, the standard stochastic gradient descent updates \cite{recht10a, teflioudi12a},
\begin{equation}\label{eq:Euclidean_stochastic}
\begin{array}{lllll}
{\mat{L}_b}_+  =   {\mat{L}_b}   -   t \mat{S}_b \mat{R}_b\\
{\mat{R}_b}_+   =  {\mat{R}_b}  -  t \mat{S}_b^T \mat{L}_b,\\
\end{array}
\end{equation} 
are not scale invariant. Here $\mat{L}_b$ and $\mat{R}_b$ are the submatrices of $\mat L$ and $\mat R$, respectively that are updated for the batch size $b$, $t$ is the stepsize, and $\mat{S}_b$ is the residual matrix at $(\mat{L}_b, \mat{R}_b)$. 

The benefits of scale invariant algorithms are discussed in \cite{absil08a, meyer11a, mishra16a}. We also show an example in Section \ref{sec:numerical_comparisons_scale_invariance}-B.

\textbf{C. Choice of $\mu$.}
A key observation is that the updates (\ref{eq:modified_scaled_stochastic}) with $\mu = 1$ are equivalent to the stochastic version of the updates proposed in \cite{mishra12a, tanner15a, wen12a}. \changeBM{For $\mu < 1$, we take the additional local information into account.} On the other hand, $\mu = 0$ gives full weighting to the ``local'' second order information and should be used when $\mat{L}_b^T\mat{L}_b$ and $\mat{R}_b^T\mat{R}_b$ are positive definite. This holds true for $b_{\mat{L}} \geq  r$ and $b_{\mat{R}} \geq  r$, i.e., for a large enough batch size. For a smaller batch size, i.e., $b_{\mat{L}} < r$ or $b_{\mat{R}} < r$, a non-zero $\mu$ should be used. In problem instances where a large number of entries are already known, i.e., $|\Omega|$ is large, the influence of $\mu < 1$ is minimal. However, for \emph{ill-conditioned data}, making use of local information is more critical, and a smaller value of $\mu$ is more appropriate, e.g., $\mu = 0.5$. These trade-offs are shown in Section \ref{sec:numerical_comparisons_mu}-C.

\textbf{D. Choice of batch size $b$.}
For $b = |\Omega|$, the algorithm in Table \ref{tab:Riemannian_stochastic_gradient_descent} behaves like a batch gradient descent algorithm and with same computational cost as discussed Section \ref{sec:computational_cost}-A. The choice of $b = 1$ is more appropriate for a fully online system. Other choices depend on the problem size and set up. Section \ref{sec:influence_b}-D shows the robust behavior of the algorithm with different choices of $b$.

\textbf{E. Convergence.}
The convergence analysis of the proposed algorithm in Table \ref{tab:Riemannian_stochastic_gradient_descent} follows the discussion in \cite{teflioudi12a, recht13a, bottou98a} except for the (positive definite) matrix-scaling of the partial derivatives $(\mat{S}_b \mat{R}_b,  \mat{S}_b^T \mat{L}_b)$ at $(\mat{L}_b, \mat{R}_b)$ by multiplying with $((\frac{b\mu}{\max(m, n)}(\mat{R}^T\mat{R}) + (1 -\mu) (\mat{R}_b^T\mat{R}_b) ) ^{-1}, \allowbreak (\frac{b\mu}{\max(m, n)}(\mat{L}^T\mat{L}) + (1-\mu) (\mat{L}_b^T\mat{L}_b) ) ^{-1})$. 

A different interpretation is that we endow the search space $\mathbb{R}^{n \times r}\times \mathbb{R} ^{m \times r}$ with the \emph{adaptive inner product} 
\begin{equation}\label{eq:modified_metric}
\begin{array}{lll}
 \langle  {  \xi}_{{x}} ,  {\eta}_{{x}}  \rangle_{\rm adaptive} = \\
   \trace  ((\frac{b\mu}{\max(m, n)}(\mat{R}^T\mat{R}) + (1 -\mu) (\mat{R}_b^T\mat{R}_b) ){\xi}_{\mat L}^T  {\eta}_{\mat L}) \\
      + \trace ((\frac{b\mu}{\max(m, n)}(\mat{L}^T\mat{L}) + (1-\mu) (\mat{L}_b^T\mat{L}_b) ) {\xi}^T_{\mat R}{\eta}_{\mat R}  ), \\
\end{array}
\end{equation}
which depends on $(\mat{L}, \mat{R})$ and $(\mat{L}_b, \mat{R}_b)$. Here ${x}$ has the matrix representation $(\mat{L}, \mat{R})$ and ${\xi}_{{x}} $ and ${\eta}_{{x}} $ vectors in $\mathbb{R}^{n \times r} \times \mathbb{R}^{m \times r}$ with matrix representations $({\xi}_{\mat L}, {\xi}_{\mat R})$ and  $({\eta}_{\mat L}, {\eta}_{\mat R})$, respectively. $\langle {\xi}_{{x}}, {\eta}_{{x}} \rangle _{\rm adaptive}$ is the inner product between ${\xi}_{{x}} $ and ${\eta}_{{x}} $ and $\trace (\cdot)$ is the matrix trace operator. Finally, computing the steepest descent directions with the inner product (\ref{eq:modified_metric}) leads to the updates (\ref{eq:modified_scaled_stochastic}). The proposed inner product (\ref{eq:modified_metric}) has the interpretation of a \emph{Riemannian metric} in the framework of Riemannian optimization \cite{absil08a}. Consequently, the analysis of a Riemannian stochastic gradient descent algorithm presented in \cite{ bonnabel13a} is applicable to our case, e.g., under a decaying stepsize condition and in a compact region near the minimum.

While asymptotic convergence guarantees for the proposed algorithm are available, the rate of convergence analysis of the proposed algorithm in Table \ref{tab:Riemannian_stochastic_gradient_descent} is a challenging task and remains an open problem. We point to the recent work \cite{sa2015} that sheds more light on this for a similar problem.

\section{Numerical experiments}\label{sec:numerical_comparisons}
In this section, we compare the following algorithms. 

(1) \textbf{Scaled-SGD:} the proposed scaled stochastic gradient descent algorithm. The computation cost is $O(|\Omega| (r^3/ b + b_{\mat L}r^2/b +  b_{\mat R}r^2/b  + r + {\rm log}b) )$ after we have seen all the $| \Omega|$ entries with $b$ entries at a time. $\mat{L}^T\mat{L}$ and $\mat{R}^T\mat{R}$ are updated after every update. (2) \textbf{SGD:} the standard SGD implementation, where the updates result by imposing the Euclidean metric \cite{recht10a, teflioudi12a}. The computational cost is $O(|\Omega|(r + {\rm log}b) )$ when we sweep through $|\Omega|$ entries with $b$ entries at a time. It should be stated that the cost is independent of $b$ as the effect of ${\rm log}b$ is minimal. (3) \textbf{Grouse:} the algorithm proposed in \cite{balzano10a}. Instead of learning $\mat{L}$ and $\mat{R}$ simultaneously, it learns, e.g., the rank-$r$ left subspace spanned by $\mat{L}$ first by traversing through the columns of the incomplete matrix $\mat{X}^\star$. The $\mat{R}$ factor is computed by solving a least-squares problem in closed form once we have learned the subspace spanned by $\mat L$. The learning of $\mat{L}$ boils down to an optimization problem on the Grassmann manifold. Consequently, Grouse is a stochastic gradient descent algorithm on the Grassmann manifold. Its computational cost is $O(mnr)$ after sweeping through $m$ columns, i.e., one pass through $\mat{X}^\star$. (4) \textbf{Loreta:} the algorithm proposed in \cite{shalit10a}, but modified to handle data as in \cite{balzano10a}, i.e., we sweep through the incomplete matrix $\mat{X}^\star$ column by column. The computational cost is $O(mnr)$ after one pass through $\mat{X}^\star$. (5) \textbf{ALS:} the standard alternating least-squares algorithm, where we update the low-rank factors $\mat{L}$ and $\mat{R}$ row-by-row by solving the least-squares subproblems in closed form \cite{zhou08a}. The computational cost is $O(|\Omega| r^2 +  (n + m)r^3)$ per update all the rows of $\mat{L}$ and $\mat R$. (6) \textbf{CCD++:} the algorithm proposed in \cite{yu14a}, which learns $r$ rank-1 factors sequentially. For learning a rank-1 factor, it uses $\rm T$ inner iterations of the algorithm proposed in \cite{pilaszy10a}. Its computational cost is $O(|\Omega| r{\rm T})$ after one update of $\mat{L}$ and $\mat R$. As suggested in \cite{yu14a}, $\rm T$ is set to $5$.

The choice of the above algorithms is motivated by the fact that these algorithms can be readily adapted to an online setup. As the mentioned algorithms are well suited for different scenarios and have implementations in different programming languages, we use only their Matlab implementations (which we implement for all except Loreta and Grouse) and compare them on the behavior of the cost function against \emph{iterations}. An iteration for Scaled-SGD, SGD, Grouse, and Loreta corresponds to one pass through $|\Omega|$ entries. For ALS and CCD++, an iteration corresponds to one update of the low-rank factors $\mat{L}$ and $\mat R$. During each iteration, the stepsize $t$ is fixed for Scaled-SGD, SGD, and Loreta. The stepsize is then updated according to the bold driver heuristic as suggested in \cite{teflioudi12a}. In the bold driver protocol \cite{battiti89a}, updating of the stepsize depends on the cost function. In case the cost increases after an iteration, the stepsize is reduced by $50\%$, else the stepsize is increased by $10\%$. The initial stepsize is computed using the approach by linearizing the cost function as done in \cite{mishra12a}. For Grouse, we use the stepsize update proposed in \cite{balzano10a}. ALS and CCD++ do not require any stepsize tuning. Additionally, during each iteration, the known entries are randomly chosen for Scaled-SGD and SGD with \emph{uniform} probability and without replacement. Equivalently, each known entry is seen only once per iteration. Similarly, during an iteration, the columns (rows) are randomly and uniformly chosen for Grouse and Loreta (ALS) without replacement.

\textbf{A. Experimentation setup and stopping criteria.} All simulations are performed in Matlab and on a $2.7$ GHz Intel Core $\rm{i}5$ machine with $8$ GB of RAM. For each example, an $n\times m$ random matrix of rank $r$  is generated as in \cite{cai10a}. Two matrices $\mat{A} \in \mathbb{R}^{ n\times r}$ and $\mat{B} \in \mathbb{R}^{ m\times r}$ are generated according to a Gaussian distribution with zero mean and unit standard deviation. The matrix product $\mat{AB}^T$ gives a random rank-$r$ matrix. A fraction of the entries are randomly removed with uniform probability. The dimensions of $n \times m$ matrices of rank $r$ is $(n + m - r)r$. The over-sampling (OS) ratio determines the number of entries that are known. An $\OS $ of $6$ implies that $6(n + m - r)r$ number of randomly and uniformly selected entries are known a priori out of the total $nm$ entries. No regularization is used. The algorithms are stopped when either the mean square error $\|\mathcal{P}_{\Omega}(\mat{X}) - \mathcal{P}_{\Omega}(\mat{X^\star})\|_F^2 /|\Omega|$ is less than $10^{-8}$ or the relative residual $\|\mathcal{P}_{\Omega}(\mat{X}) - \mathcal{P}_{\Omega}(\mat{X^\star})\|_F / \|\mathcal{P}_{\Omega}(\mat{X^\star})\|_F$ is less than $10^{-4}$ or the number of iterations exceeds $100$. The Matlab codes are available at \url{http://bamdevmishra.com/codes/scaledSGD/}.

\begin{figure*}[t]
\begin{tabular}{cc}
\hspace*{0.8cm}
\noindent \begin{minipage}[b]{0.26\hsize}
\centering
\includegraphics[width=\hsize]{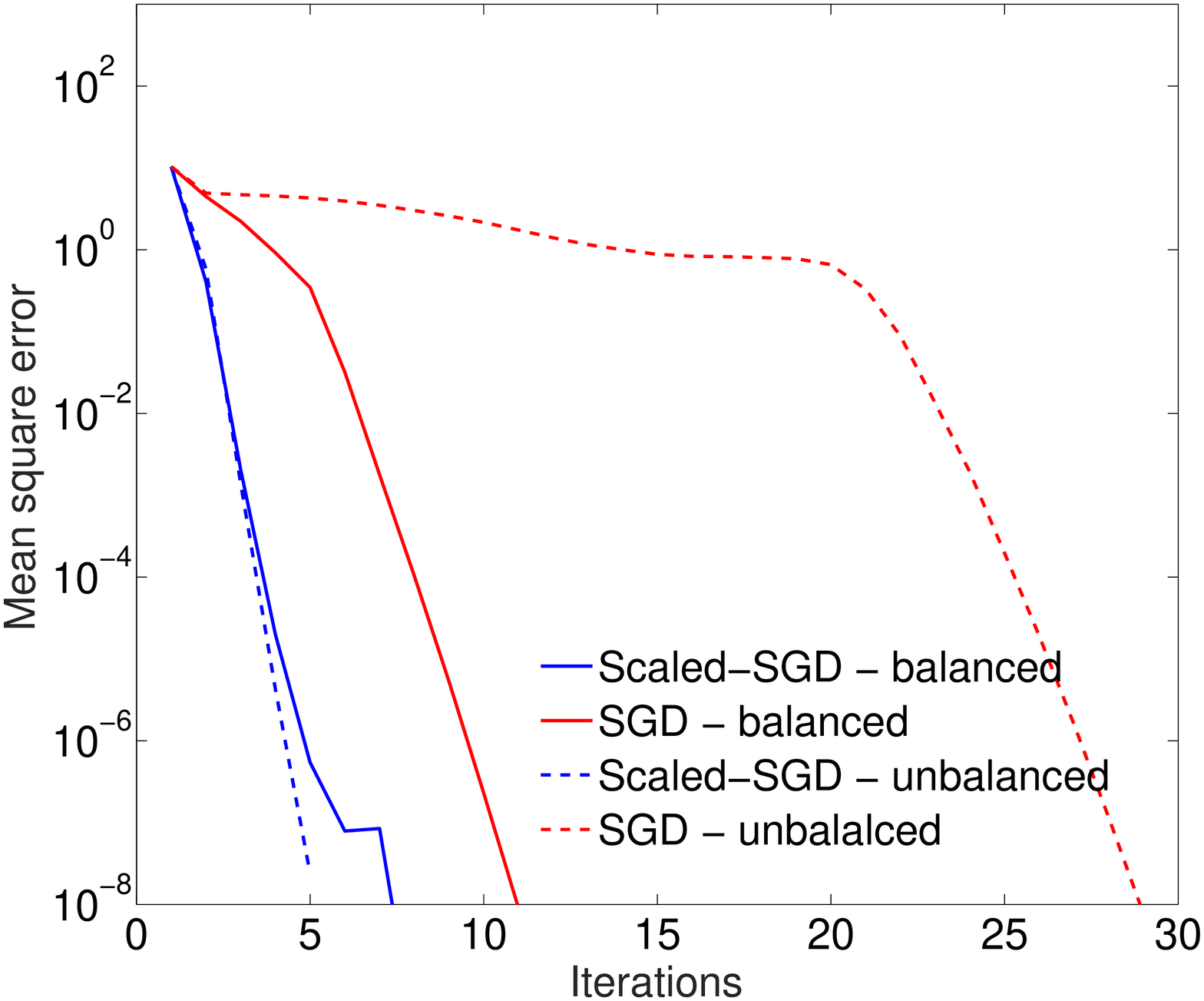}
{\scriptsize (a) Effect of scale invariance.}
 \end{minipage}
 \noindent \begin{minipage}[b]{0.26\hsize}
\centering
\includegraphics[width=\hsize]{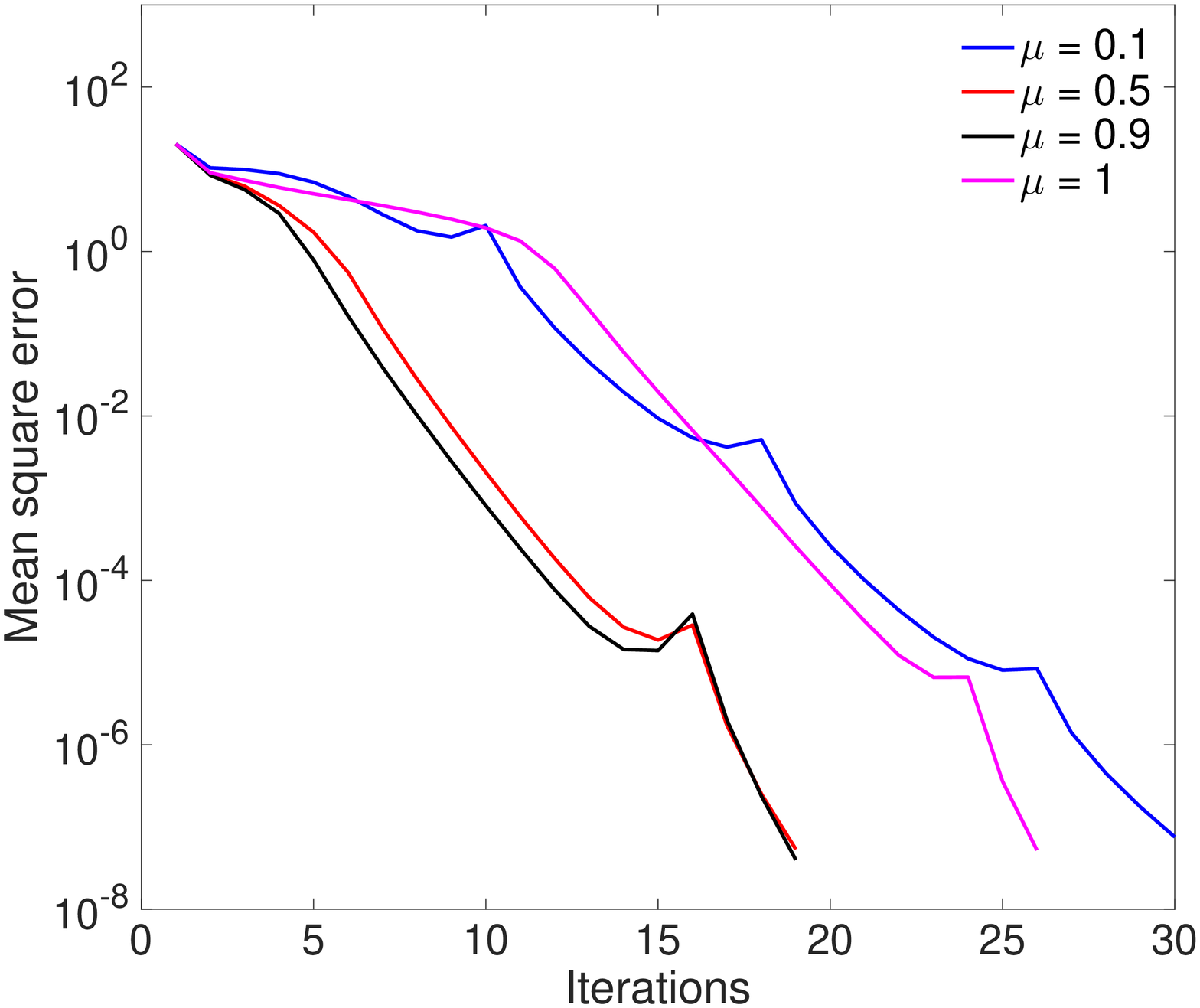}
{\scriptsize (b) Effect of $\mu$ on well-conditioned data.}
 \end{minipage}
\noindent \begin{minipage}[b]{0.26\hsize}
\centering
\includegraphics[width=\hsize]{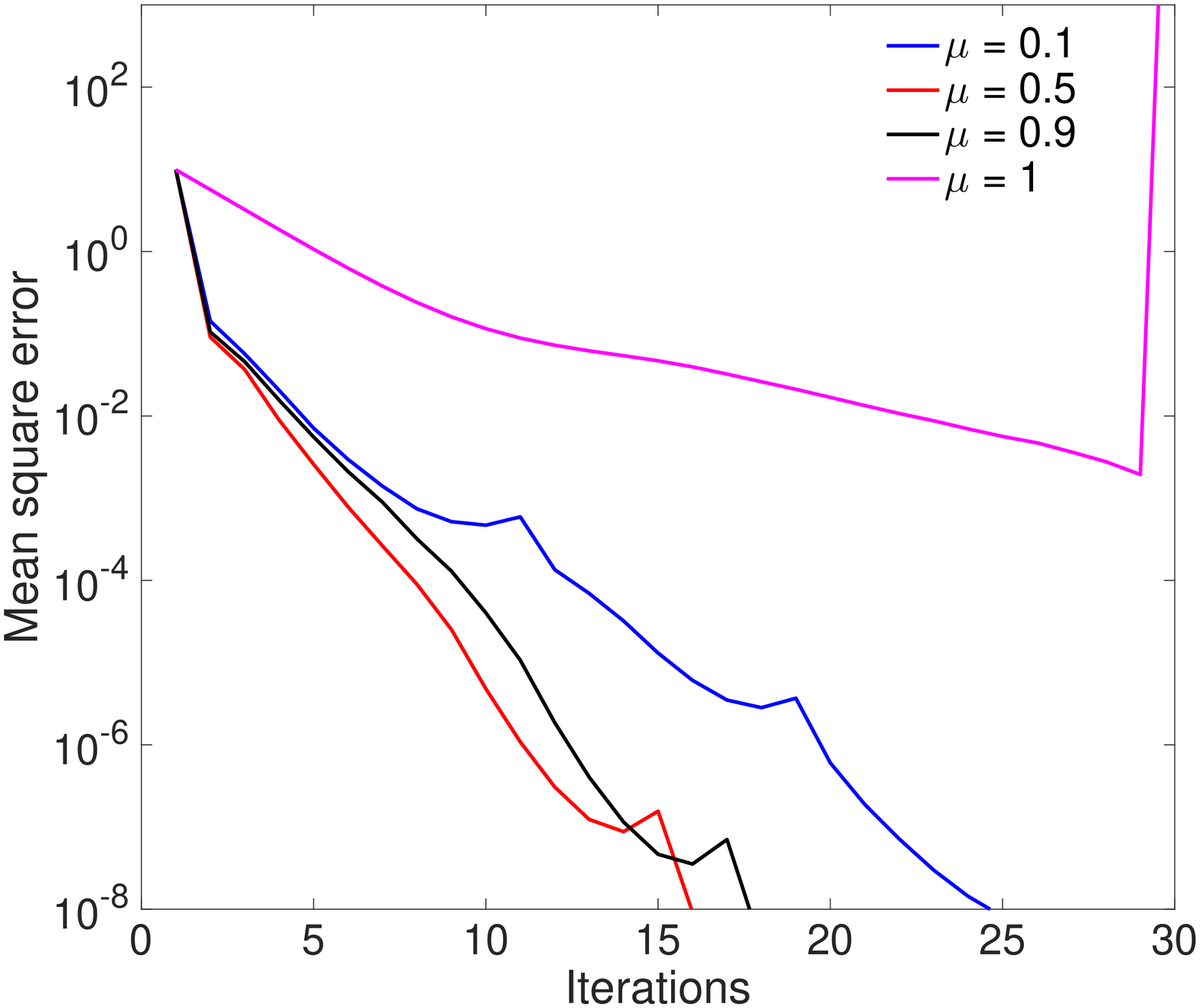}
{\scriptsize (c) Effect of $\mu$ on data with ${\rm CN} = 50$.}
 \end{minipage}\\
 \hspace*{0.8cm}
\noindent \begin{minipage}[b]{0.26\hsize}
\centering
\includegraphics[width=\hsize]{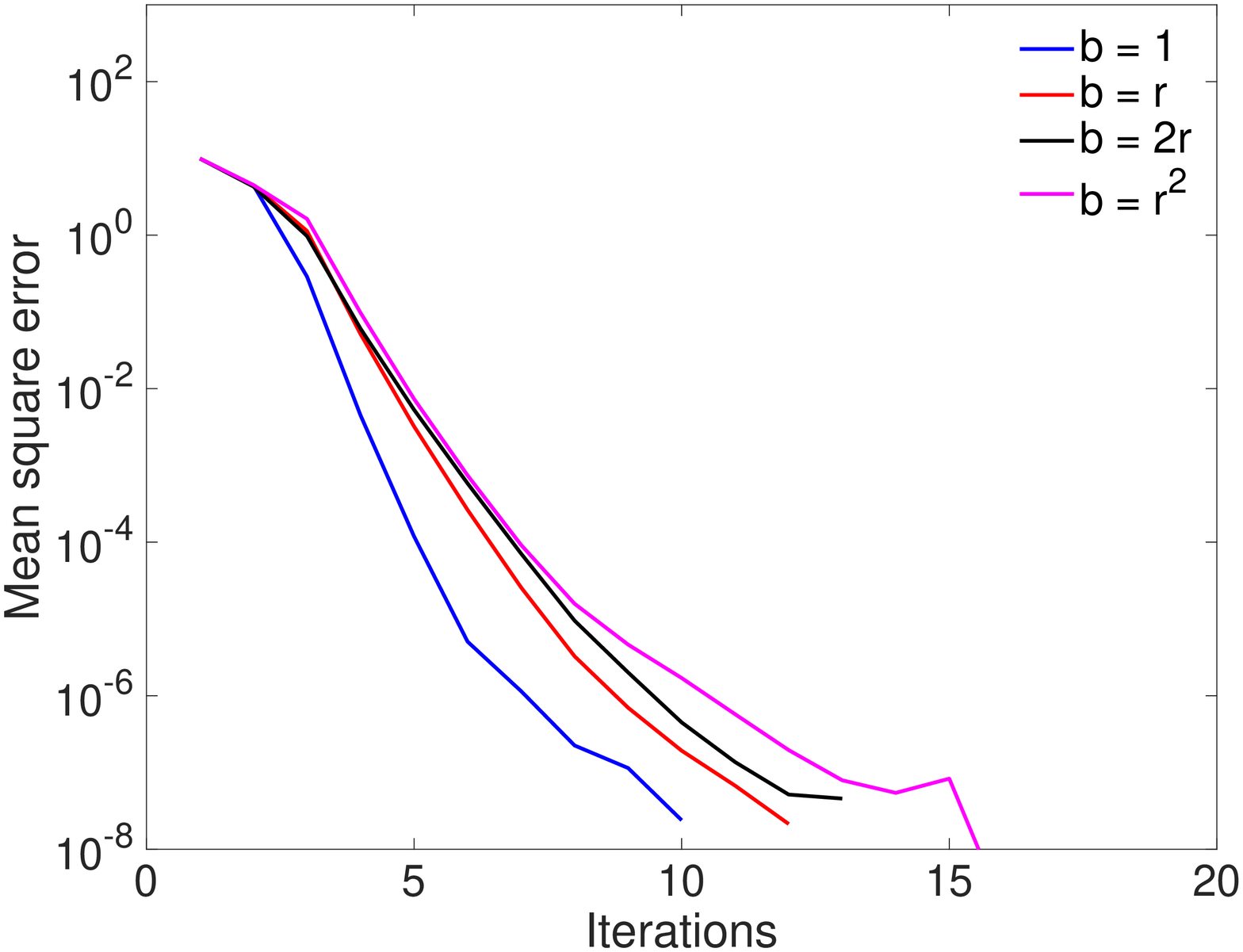}
{\scriptsize (d) Effect of $b$ on well-conditioned data.}
 \end{minipage}
\noindent \begin{minipage}[b]{0.26\hsize}
\centering
\includegraphics[width=\hsize]{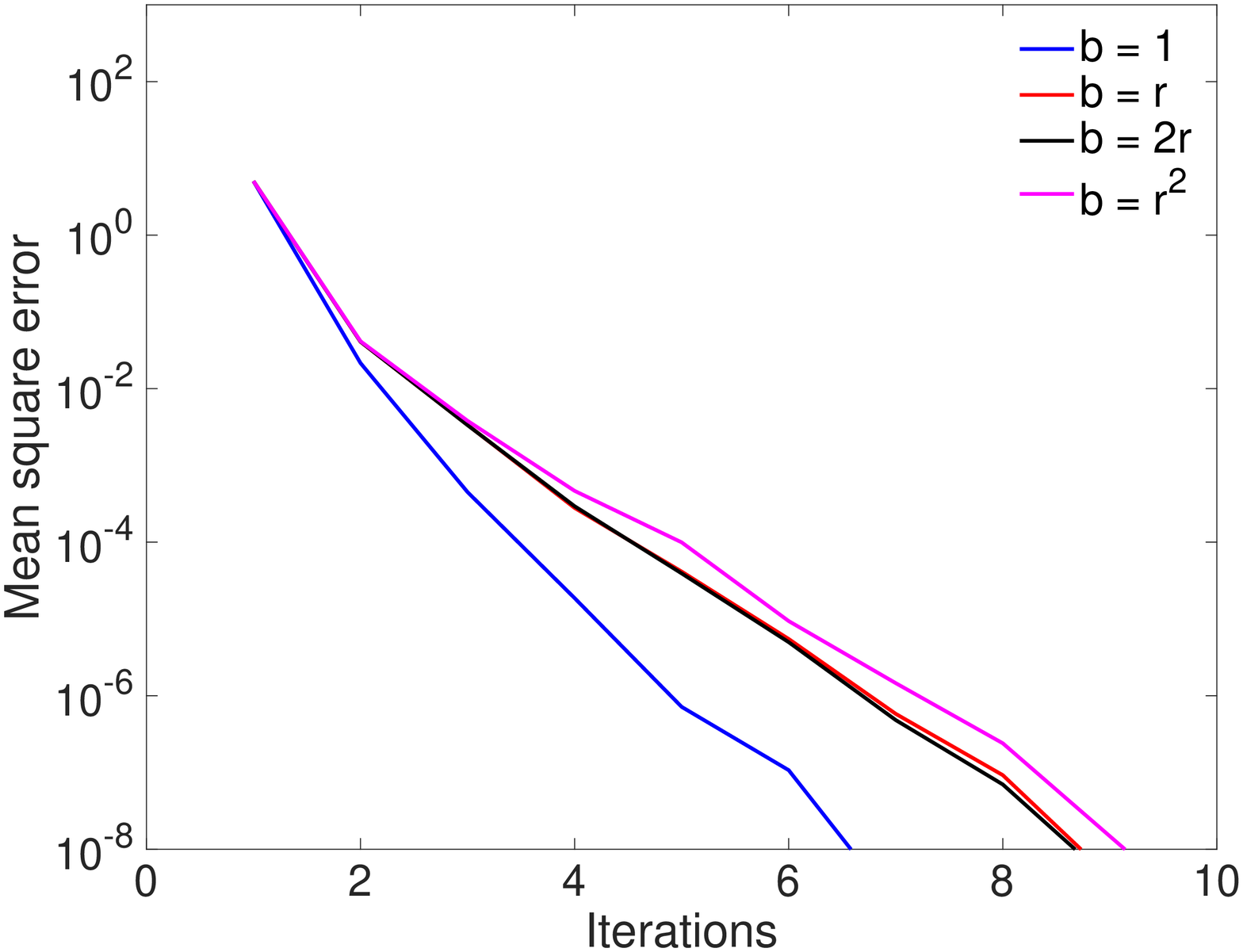}
{\scriptsize (e) Effect of $b$ on data with ${\rm CN} = 500$.}
\end{minipage}
\noindent \begin{minipage}[b]{0.26\hsize}
\centering
\includegraphics[width=\hsize]{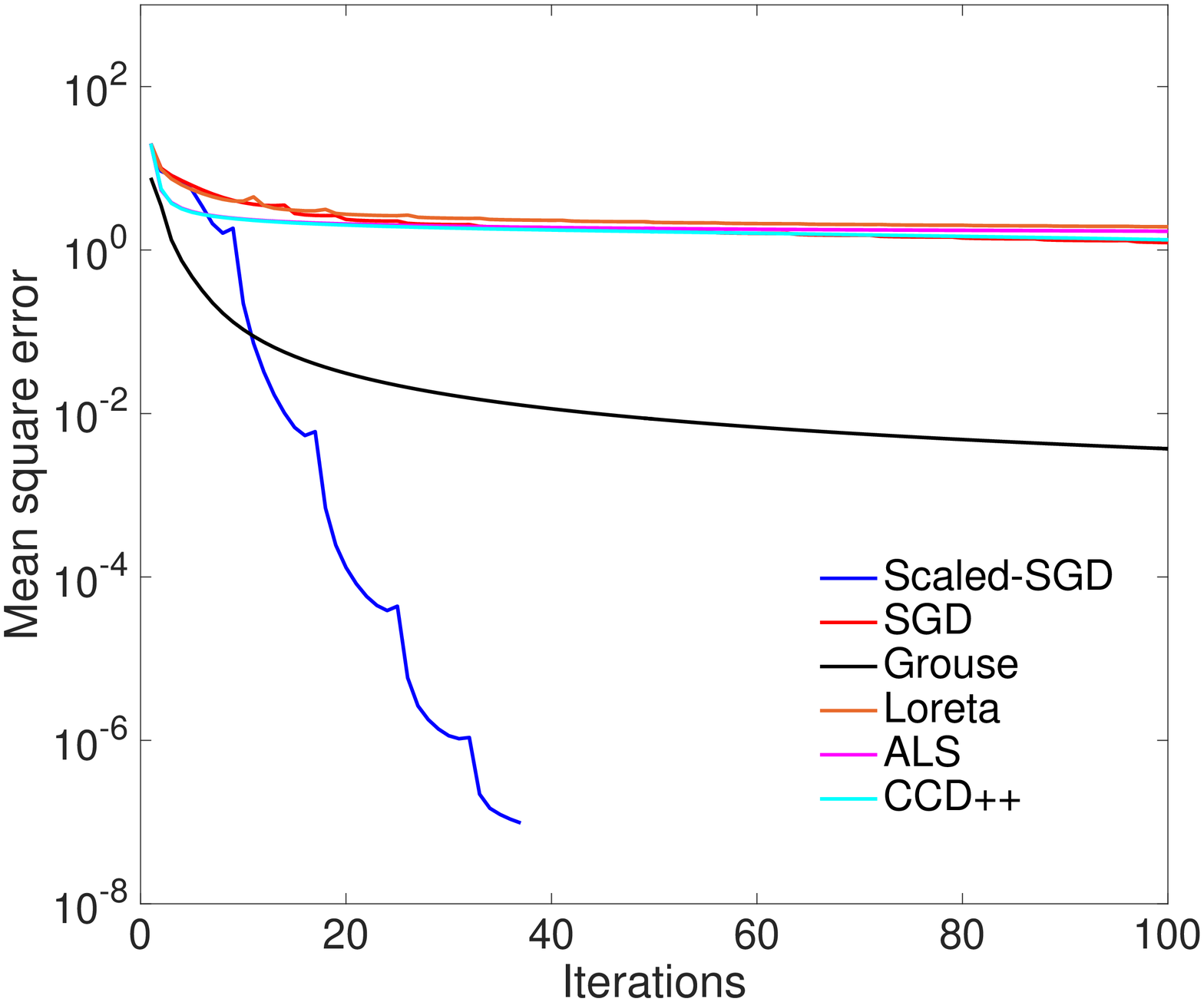}
{\scriptsize (f) Low-sampling with ${\rm OS} = 2.1$.}
 \end{minipage}\\
 \hspace*{0.8cm}
\noindent \begin{minipage}[b]{0.26\hsize}
\centering
\includegraphics[width=\hsize]{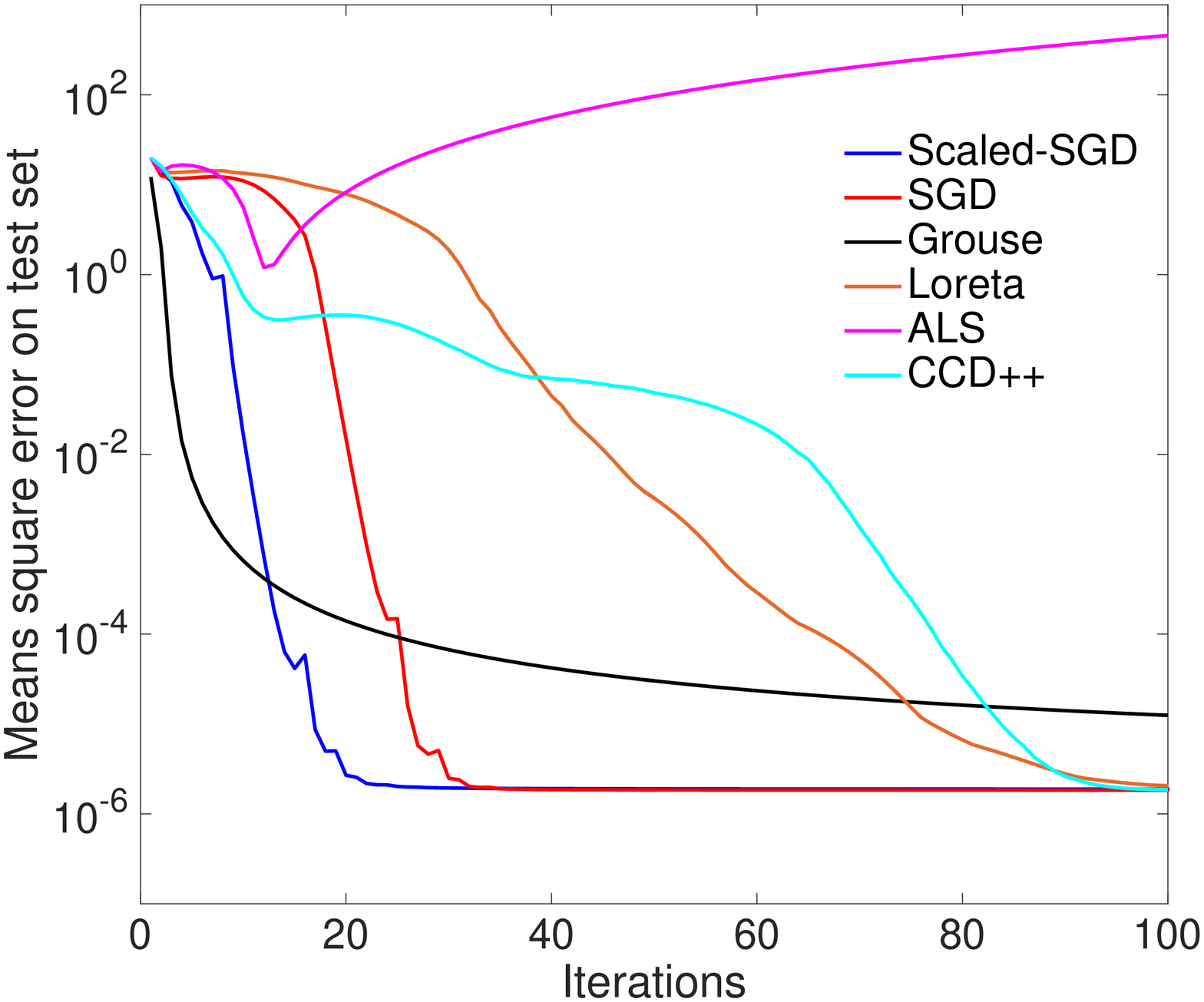}
{\scriptsize (g) Noisy instance.}
 \end{minipage}
\noindent \begin{minipage}[b]{0.26\hsize}
\centering
\includegraphics[width=\hsize]{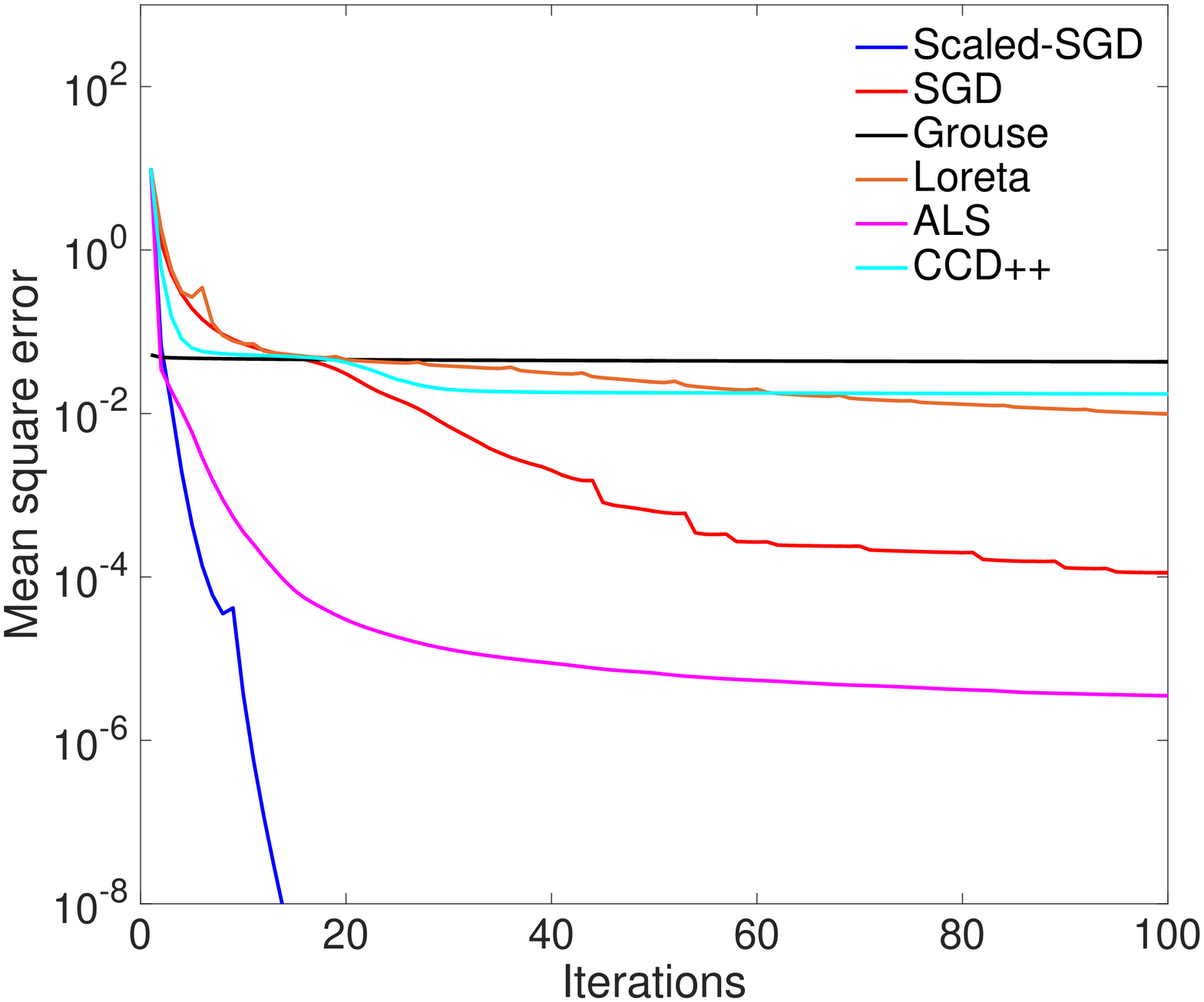}
{\scriptsize (h) Ill-conditioned data with ${\rm CN} = 100$.}
 \end{minipage}
\noindent \begin{minipage}[b]{0.26\hsize}
\centering
\includegraphics[width=\hsize]{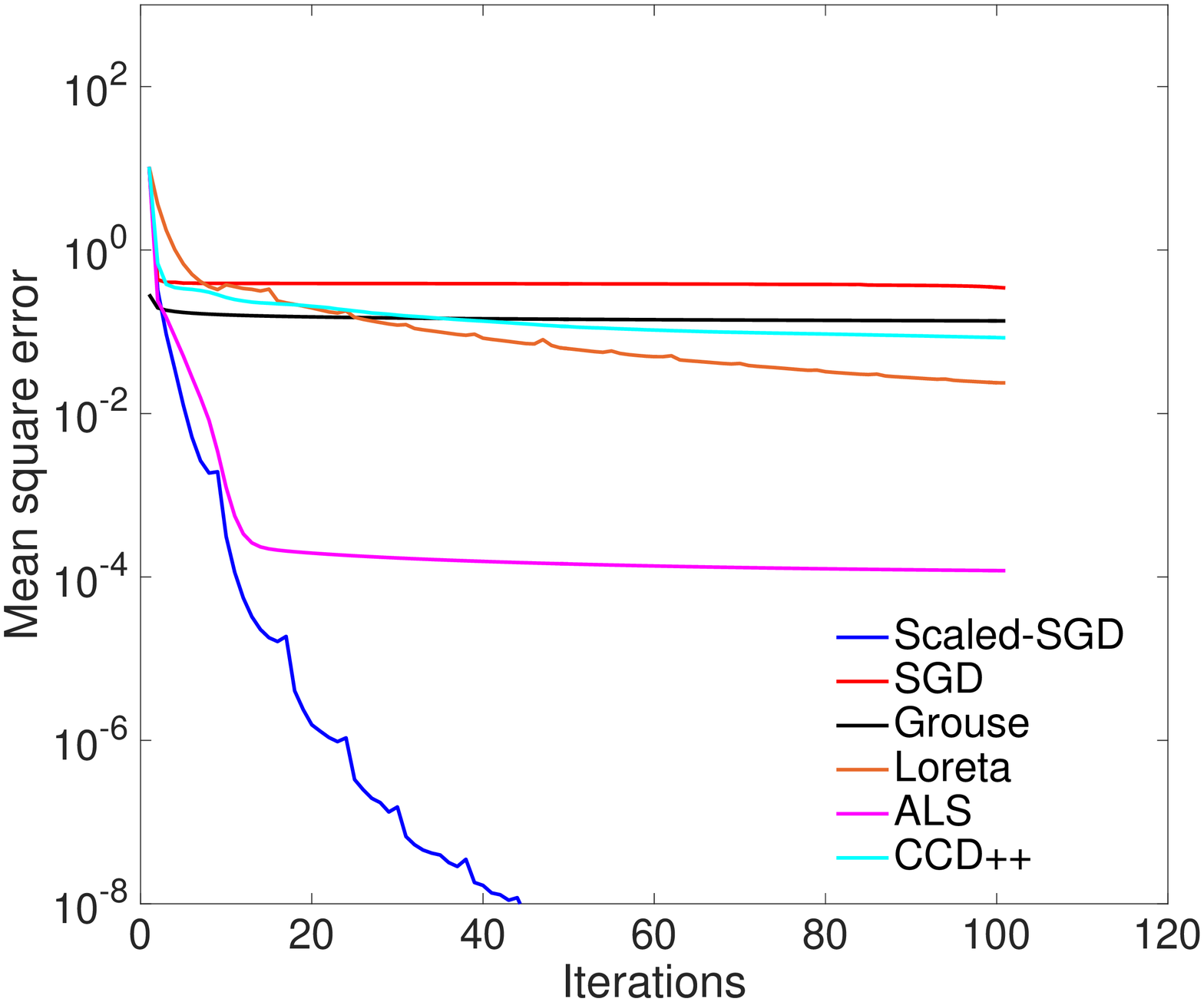}
{\scriptsize (i) Rectangular instance.}
 \end{minipage}
 \end{tabular}
\caption{Scaled-SGD resolves scale-invariance in matrix factorization. Furthermore, it is robust to various choices of $\mu$ and $b$. In particular, a good trade-off between global and local information, i.e., for $\mu \in (0,1)$, leads to improved performance in both well-conditioned and ill-conditioned data. Scaled-SGD outperforms others in lower sampling instances, i.e., with small ${\rm OS}$ values and ill-conditioned instances.}
\label{fig:all}
\end{figure*}

\textbf{B. Effect of scale invariance.} \label{sec:numerical_comparisons_scale_invariance}
The difference of the standard stochastic updates (\ref{eq:Euclidean_stochastic}) with the proposed updates (\ref{eq:modified_scaled_stochastic}) is in the $r\times r$ matrices, e.g., $(\frac{b\mu}{\max(m, n)}(\mat{R}^T\mat{R}) + (1 -\mu) (\mat{R}_b^T\mat{R}_b) )$, that are inversely applied to (\ref{eq:Euclidean_stochastic}). However, those extra (but minimal) computations make the proposed updates \emph{invariant} to the transformation (\ref{eq:symmetry}), which is not the case with the Euclidean updates (\ref{eq:Euclidean_stochastic}). To illustrate this effect of scale invariance, we consider a problem instance with $ n = m= 100$, $r = 5$, and $\OS=8$. Both Scaled-SGD and SGD are run with batch size $b = 10$ and $\mat L$ and $\mat R$ are randomly initialized \changeBM{with balanced factors} (solid line in Figure \ref{fig:all}(a)) such that $\|\mat{L}_{\rm init}\|_F \approx \|\mat{R}_{\rm init}\|_F$. Additionally, we set $\mu$ to $0.5$. The performance of both the algorithms is similar. However, the performance of SGD suffers drastically for \changeBM{unbalanced factors (dashed line), i.e.,} when $\|\mat{L}_{\rm init}\|_F \approx 4\|\mat{R}_{\rm init}\|_F$ as shown in Figure \ref{fig:all}(a).


\textbf{C. Effect of $\mu$.} \label{sec:numerical_comparisons_mu}
We consider problem instances of size $5000\times 5000$ and rank $10$. In order to understand the influence of $\mu$ on Scaled-SGD, we consider two scenarios. The first scenario consists of an instance with over-sampling ratio $3$. The second scenario considers an over-sampling ratio of $3$ and ill-conditioned data with condition numbers (CN) $50$, which is obtained by imposing an exponential decay of singular values (discussed in Section \ref{sec:numerical_comparisons_ill_conditioned}-G). Figures \ref{fig:all}(b) and \ref{fig:all}(c) show the behavior of Scaled-SGD with four different values of $\mu$. Scaled-SGD is run with batch size $b = 10$. Figures \ref{fig:all}(b) shows that there exists values of $\mu$, which show better performance. Figure \ref{fig:all}(c) shows that relying solely \changeBM{on} global information, i.e., $\mu = 1$, need not be better. In particular, Scaled-SGD with $\mu = 1$ diverges in \ref{fig:all}(c). $\mu = 0.5$, on the other hand, shows a good performance in many instances.

\textbf{D. Effect of batch size $b$.}\label{sec:influence_b}
We consider problem instances of size $5000\times 5000$ of rank $5$. In order to understand the influence of $b$ on Scaled-SGD, we consider two scenarios with $b = \{ 1,r, 2r, r^2\}$ \changeBM{with over-sampling ratio of $5$}. The first scenario consists of \changeBM{a well-conditioned} instance. The second scenario consists of ill-conditioned data with condition number ${\rm CN}$ equal to $500$. $\mu$ is set to $0.5$ in Scaled-SGD. Figures \ref{fig:all}(d) and \ref{fig:all}(e) show the robust behavior of Scaled-SGD with different batch sizes $b$.

\textbf{E. Low-sampling instances.}\label{sec:low_sampling}
We consider problem instances of size $5000\times 5000$ and rank $10$. Different over-sampling ratios of $4$, $3$, $2.5$, and $2.1$ are considered. Scaled-SGD and SGD are run with batch size $b = 10$. Additionally, we set $\mu$ to $ 0.5$ in Scaled-SGD. While most algorithms perform well for larger OS values, Scaled-SGD particularly outperforms others for smaller OS values as shown in Figure \ref{fig:all}(f).


\textbf{F. Noisy instances.} We consider the problem instance in Section \ref{sec:low_sampling}-E with $\OS = 3$. Additionally, noise is added to the known entries. As proposed in \cite{balzano10a}, noise for each entry is sampled from the Gaussian distribution with mean zero and standard deviations $10^{-4} $. Figure \ref{fig:all}(g) shows the performance of the algorithms on the test set that is held out, which is different from the training set $\Omega$.


\textbf{G. Ill-conditioned instances.}\label{sec:numerical_comparisons_ill_conditioned}
We consider matrices of size $5000 \times  5000$ and rank $10$ and impose an exponential decay of singular values. The ratio of the largest to the lowest singular value is known as the condition number (CN) of the matrix. For example, at rank $10$ the singular values with condition number $100$ is obtained using the Matlab function \texttt{logspace(-2,0,10)}. The over-sampling ratio for these instances is $3$. $\mu$ is set to $ 0.5$. Figure \ref{fig:all}(h) shows the performance of various algorithms, where our proposed approach outperforms others for $\rm{CN}$ $100$. Scaled-SGD shows a robust performance on ill-conditioned instances.


\textbf{H. Rectangular instances.}
We consider rectangular matrices of size $1000\times 8000$ of rank $10$ and over sampling ratio $3$. $\mu$ is set to $ 0.5$. Most comparisons suggest that ALS and Grouse perform very well on those instances. However, even for slightly ill-conditioned data, the performance of ALS and Grouse degrade as shown in Figure \ref{fig:all}(i). Scaled-SGD remains unaffected.


\begin{figure*}[htbp]
\parbox[t]{6.5cm}{\null
  \centering
 \includegraphics[width=0.8\hsize]{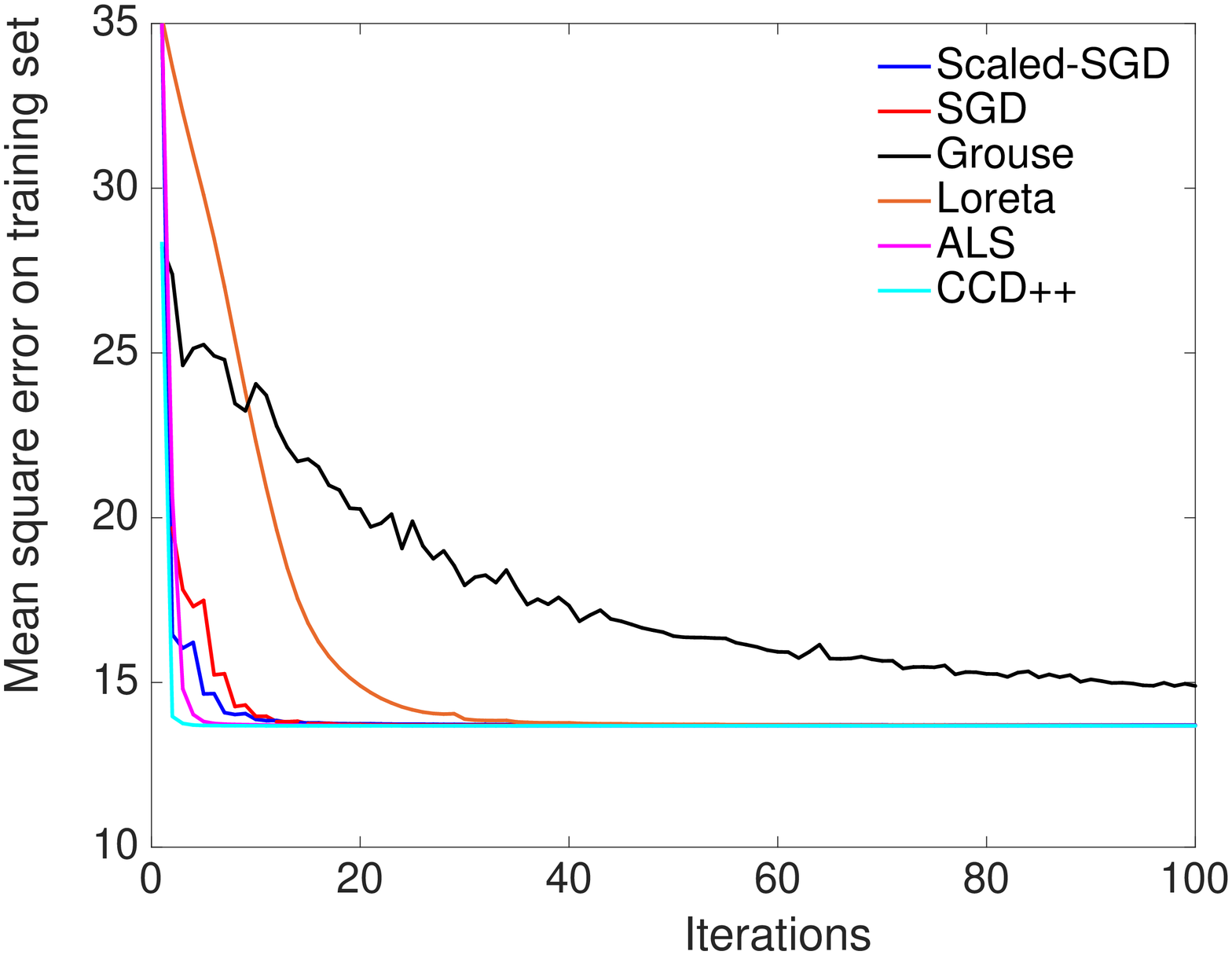}
  \captionof{figure}{{Performance of the algorithms on the Jester dataset for $n = 2000$ and $r = 7$.}} \label{fig:jester}
}
\parbox[t]{11cm}{\null
\centering
 \vskip-\abovecaptionskip
 \captionof{table}[t]{{Final mean NMAEs obtained on the test set of the Jester dataset.}}
 \vskip\abovecaptionskip
  {\scriptsize 
\begin{tabular}{p{1.5cm}|p{1.0cm}|p{1.0cm}|p{1.cm}|p{1.0cm}|p{1.0cm}|p{1.0cm}}
\hlinewd{1.0pt}
\textbf{Algorithm}& \multicolumn{2}{c|}{$n = 2000$}  & \multicolumn{2}{c|}{$n = 5000$} & \multicolumn{2}{c}{$n = 24983$}  \\
\hdashline
& $r = 5$ &$r = 7$ &$r = 5$ & $r = 7$& $r = 5$ & $r = 7$ \\
\hline 
Scaled-SGD &$0.158$ & $0.159$ & $0.160$ & $0.158$ & 0.159 & 0.157\\
\hdashline
SGD & $0.158$ & $0.159$ & $0.160$ & $0.158$ & 0.159 & 0.157\\
\hdashline
Grouse & $0.165$ & $0.165 $ & $ 0.179$ & $0.177$ & 0.168 & 0.166\\
\hdashline
Loreta & $0.159$ & $0.160$ & $0.160$ & $0.158$ &0.159 & 0.158\\
\hdashline
ALS& $0.158$ & $0.159$ & $0.160$ & $0.158$ & 0.159 & 0.157\\
\hdashline
CCD++ & $0.158$ & $0.159 $ & $0.160$ & $0.157$ & 0.159 & 0.157\\
\hlinewd{1.0pt}
\end{tabular}
\label{tab:jester}
}
}
\end{figure*}

\textbf{I. Jester dataset.} 
We consider the Jester dataset 1 \cite{goldberg01a} consisting of ratings of $100$ jokes by $24 983$ users. Each rating is between $-10$ and $10$. Following the protocol in \cite{keshavan09a}, we select $n = \{2000, 5000, 24983 \}$ users randomly. We randomly extract two ratings per user as test data. The algorithms are run for ranks $\{5,7\}$ with random initialization and for $100$ iterations. Predictions are computed at the end of $100{\rm th}$ iteration. The entire process is repeated ten times. No regularization is used for Scaled-SGD, SGD, Grouse, and Loreta. The performance of ALS and CCD++, however, \emph{critically} depends on regularization for which we set the regularization parameter to $10$ for $n = \{2000, 5000 \}$ and $100$ for $n = 24983$ after cross-validation. The batch size $b$ is set to $r$ for Scaled-SGD and SGD. $\mu$ is set to $0.5$ in Scaled-SGD. Figure \ref{fig:jester} shows the performance plots. Table \ref{tab:jester} shows the \emph{final} normalized mean absolute errors (NMAE) obtained by different algorithms on the test dataset averaged over \emph{ten} runs. NMAE is defined as the mean absolute error (MAE) divided by spread of the ratings, i.e., the difference between the minimum and maximum ratings, i.e., NMAE is MAE/$20$. The standard deviation of the scores in Table \ref{tab:jester} is $2\cdot 10^{-3}$. Except Grouse, all other algorithms give similar NMAE scores on the test set. Scaled-SGD consistently outperforms SGD as shown in Figure \ref{fig:jester}.


\section{Conclusion}
We have proposed a scaled variant of stochastic gradient descent algorithm for the low-rank matrix completion problem. It is based on a novel matrix-scaling of the partial derivatives with terms that combine both local and global second order information. This scaling is computationally cheap to implement and the proposed algorithm is potentially scalable to larger datasets. Initial results show a robust performance of the proposed algorithm on various benchmarks. At the conceptual level, this paper shows the complementary role of local and global second order information in a stochastic gradient setting.

\bibliography{scaled_sgd}

\begin{thebibliography}{10}
\providecommand{\url}[1]{#1}
\csname url@samestyle\endcsname
\providecommand{\newblock}{\relax}
\providecommand{\bibinfo}[2]{#2}
\providecommand{\BIBentrySTDinterwordspacing}{\spaceskip=0pt\relax}
\providecommand{\BIBentryALTinterwordstretchfactor}{4}
\providecommand{\BIBentryALTinterwordspacing}{\spaceskip=\fontdimen2\font plus
\BIBentryALTinterwordstretchfactor\fontdimen3\font minus
  \fontdimen4\font\relax}
\providecommand{\BIBforeignlanguage}[2]{{%
\expandafter\ifx\csname l@#1\endcsname\relax
\typeout{** WARNING: IEEEtran.bst: No hyphenation pattern has been}%
\typeout{** loaded for the language `#1'. Using the pattern for}%
\typeout{** the default language instead.}%
\else
\language=\csname l@#1\endcsname
\fi
#2}}
\providecommand{\BIBdecl}{\relax}
\BIBdecl

\bibitem{candes09b}
E.~J. Cand{\`e}s and B.~Recht, ``Exact matrix completion via convex
  optimization,'' \emph{Foundations of Computational Mathematics}, vol.~9,
  no.~6, pp. 717--772, 2009.

\bibitem{keshavan10a}
R.~H. Keshavan, A.~Montanari, and S.~Oh, ``Matrix completion from a few
  entries,'' \emph{IEEE Transactions on Information Theory}, vol.~56, no.~6,
  pp. 2980--2998, 2010.

\bibitem{wei15a}
K.~Wei, J.-F. Cai, T.~F. Chan, and S.~Leung, ``Guarantees of {Riemannian}
  optimization for low rank matrix recovery,'' arXiv preprint arXiv:1511.01562,
  Tech. Rep., 2015.

\bibitem{cai10a}
J.~F. Cai, E.~J. Cand\`es, and Z.~Shen, ``A singular value thresholding
  algorithm for matrix completion,'' \emph{SIAM Journal on Optimization},
  vol.~20, no.~4, pp. 1956--1982, 2010.

\bibitem{keshavan09a}
R.~H. Keshavan, A.~Montanari, and S.~Oh, ``Low-rank matrix completion with
  noisy observations: a quantitative comparison,'' in \emph{Annual Allerton
  Conference on Communication, Control, and Computing (Allerton)}, 2009, pp.
  1216--1222.

\bibitem{boumal15a}
N.~Boumal and P.-A. Absil, ``Low-rank matrix completion via preconditioned
  optimization on the {G}rassmann manifold,'' \emph{Linear Algebra and its
  Applications}, vol. 475, pp. 200--239, 2015.

\bibitem{recht10a}
B.~Recht, M.~Fazel, and P.~A. Parrilo, ``Guaranteed minimum-rank solutions of
  linear matrix equations via nuclear norm minimization,'' \emph{SIAM Review},
  vol.~53, no.~3, pp. 471--501, 2010.

\bibitem{yu14a}
H.-F. Yu, C.-J. Hsieh, S.~Si, and I.~S. Dhillon, ``Parallel matrix
  factorization for recommender systems,'' \emph{Knowledge and Information
  Systems}, vol.~41, no.~3, pp. 793--819, 2014.

\bibitem{tanner15a}
J.~Tanner and K.~Wei, ``Low rank matrix completion by alternating steepest
  descent methods,'' \emph{Applied and Computational Harmonic Analysis}, 2015,
  doi: http://dx.doi.org/10.1016/j.acha.2015.08.003.

\bibitem{vandereycken13a}
B.~Vandereycken, ``Low-rank matrix completion by {R}iemannian optimization,''
  \emph{SIAM Journal on Optimization}, vol.~23, no.~2, pp. 1214--1236, 2013.

\bibitem{ngo12a}
T.~T. Ngo and Y.~Saad, ``{Scaled gradients on Grassmann manifolds for matrix
  completion},'' in \emph{Advances in Neural Information Processing Systems 25
  (NIPS)}, 2012, pp. 1421--1429.

\bibitem{recht13a}
B.~Recht and C.~R{\'e}, ``Parallel stochastic gradient algorithms for
  large-scale matrix completion,'' \emph{Mathematical Programming Computation},
  vol.~5, no.~2, pp. 201--226, 2013.

\bibitem{teflioudi12a}
C.~Teflioudi, F.~Makari, and R.~Gemulla, ``Distributed matrix completion,'' in
  \emph{International Conference on Data Mining (ICDM)}, 2012, pp. 655--664.

\bibitem{shalit10a}
U.~Shalit, D.~Weinshall, and G.~Chechik, ``Online learning in the manifold of
  low-rank matrices,'' in \emph{Neural Information Processing Systems
  conference (NIPS)}, 2010, pp. 2128--2136.

\bibitem{balzano10a}
L.~Balzano, R.~Nowak, and B.~Recht, ``Online identification and tracking of
  subspaces from highly incomplete information,'' in \emph{The 48th Annual
  Allerton Conference on Communication, Control, and Computing (Allerton)},
  June 2010, pp. 704--711.

\bibitem{zhou08a}
Y.~Zhou, D.~Wilkinson, R.~Schreiber, and R.~Pan, ``Large-scale parallel
  collaborative filtering for the {Netflix} prize,'' in \emph{International
  Conference on Algorithmic Aspects in Information and Management (AAIM)},
  2008, pp. 337--348.

\bibitem{pilaszy10a}
I.~Pil\'aszy, D.~Zibriczky, and D.~Tikk, ``Fast als-based matrix factorization
  for explicit and implicit feedback datasets,'' in \emph{ACM conference on
  Recommender systems (RecSys)}, 2010, pp. 71--78.

\bibitem{wen12a}
Z.~Wen, W.~Yin, and Y.~Zhang, ``Solving a low-rank factorization model for
  matrix completion by a nonlinear successive over-relaxation algorithm,''
  \emph{Mathematical Programming Computation}, vol.~4, no.~4, pp. 333--361,
  2012.

\bibitem{markovsky13a}
I.~Markovsky and K.~Usevich, ``Structured low-rank approximation with missing
  data,'' \emph{SIAM Journal on Matrix Analysis and Applications}, vol.~34,
  no.~2, pp. 814--830, 2013.

\bibitem{shi16a}
Y.~Shi, Y.~Zhang, Letaif, and K.~B. and, ``Low-rank matrix completion for
  topological interference management by {R}iemannian pursuit,'' \emph{IEEE
  Transactions on Wireless Communications}, vol.~15, no.~7, pp. 4703--4717,
  2016.

\bibitem{meyer11a}
G.~Meyer, S.~Bonnabel, and R.~Sepulchre, ``{Linear regression under fixed-rank
  constraints: a Riemannian approach},'' in \emph{Proceedings of the 28th
  International Conference on Machine Learning (ICML)}, 2011, pp. 545--552.

\bibitem{mishra12a}
B.~Mishra, K.~Adithya~Apuroop, and R.~Sepulchre, ``A {R}iemannian geometry for
  low-rank matrix completion,'' arXiv preprint arXiv:1211.1550, Tech. Rep.,
  2012.

\bibitem{mishra16a}
B.~Mishra and R.~Sepulchre, ``{R}iemannian preconditioning,'' \emph{SIAM
  Journal on Optimization}, vol.~26, no.~1, pp. 635--660, 2016.

\bibitem{boumal14a}
N.~Boumal, B.~Mishra, P.-A. Absil, and R.~Sepulchre, ``Manopt: a {M}atlab
  toolbox for optimization on manifolds,'' \emph{Journal of Machine Learning
  Research}, vol.~15, no. Apr, pp. 1455--1459, 2014.

\bibitem{piziak99a}
R.~Piziak and P.~L. Odell, ``Full rank factorization of matrices,''
  \emph{Mathematics Magazine}, vol.~72, no.~3, pp. 193--201, June 1999.

\bibitem{battiti89a}
R.~Battiti, ``Accelerated backpropagation learning: two optimization methods,''
  \emph{Complex Systems}, vol.~3, no.~4, pp. 331--342, 1989.

\bibitem{johnson13a}
R.~Johnson and T.~Zhang, ``Accelerating stochastic gradient descent using
  predictive variance reduction,'' in \emph{Advances in Neural Information
  Processing Systems (NIPS)}, 2013, pp. 315--323.

\bibitem{absil08a}
P.-A. Absil, R.~Mahony, and R.~Sepulchre, \emph{Optimization Algorithms on
  Matrix Manifolds}.\hskip 1em plus 0.5em minus 0.4em\relax Princeton, NJ:
  Princeton University Press, 2008.

\bibitem{bottou98a}
L.~Bottou, ``Online algorithms and stochastic approximations,'' in \emph{Online
  Learning in Neural Networks}, D.~Saad, Ed.\hskip 1em plus 0.5em minus
  0.4em\relax Cambridge, UK: Cambridge University Press, 1998, pp. 9--42.

\bibitem{bonnabel13a}
S.~Bonnabel, ``Stochastic gradient descent on {R}iemannian manifolds,''
  \emph{IEEE Transactions on Automatic Control}, vol.~58, no.~9, pp.
  2217--2229, 2013.

\bibitem{sa2015}
C.~D. Sa, K.~Olukotun, and C.~R{\'e}, ``Global convergence of stochastic
  gradient descent for some non-convex matrix problems,'' in
  \emph{International Conference on Machine learning (ICML)}, 2015, pp.
  2332--2341.

\bibitem{goldberg01a}
K.~Goldberg, T.~Roeder, D.~Gupta, and C.~Perkins, ``Eigentaste: a constant time
  collaborative filtering algorithm,'' \emph{Information Retrieval}, vol.~4,
  no.~2, pp. 133--151, 2001.

\end{thebibliography}
\bibliographystyle{IEEEtran}

\end{document}